\documentclass[10pt,twocolumn,letterpaper]{article}

\usepackage[pagenumbers]{cvpr} 
\usepackage{indentfirst} 
\usepackage{amsmath,amssymb,mathtools,algorithm,algpseudocode}
\newcommand{\lfseries}{\fontseries{l}\selectfont}

\usepackage[table]{xcolor}
\usepackage{multirow}
\usepackage{graphicx}
\usepackage{booktabs}
\usepackage{enumitem}
\usepackage{caption} 
\usepackage{subcaption} 
\usepackage{amsmath}   
\usepackage{amssymb}   
\usepackage{amsfonts}  

\usepackage{amsthm}    
\theoremstyle{plain}

\theoremstyle{remark}

\numberwithin{equation}{section}  

\usepackage{tabularx}
\usepackage{makecell}

\definecolor{cvprblue}{rgb}{0.21,0.49,0.74}
\usepackage[pagebackref,breaklinks,colorlinks,allcolors=cvprblue]{hyperref}

\def\paperID{***} 
\def\confName{CVPR}
\def\confYear{2026}

\title{M\textsuperscript{4}Fuse: Lightweight State-Space MoE with a Cross-Scale Gating Bridge for Brain Tumor Segmentation
}

\author{
  Meihua Zhou$^{1}$\thanks{Equal contribution \url{github:{mh-zhou,tonxycs}/M4Fuse}}  \quad Xinyu Tong$^{1*}$\thanks{Corresponding author}
  \quad Li Yang$^{2}$\textsuperscript{\dag}  \\
  $^1$University of Chinese Academy of Sciences $^2$Wannan Medical University
 \\
{\tt\small \{zhoumeihua25, tongxinyu25\}@mails.ucas.ac.cn, yangli@wnmc.edu.cn}
}

\begin{document}
\maketitle
\begin{abstract}
Encoder-decoder imbalance and the reliance on large input volumes make many 3D brain tumor segmentation models both compute-heavy and brittle. We present M\textsuperscript{4}Fuse, a lightweight network that prioritizes discriminative brain tumor cues over exhaustive appearance reconstruction. Our method balances encoder and decoder capacity and replaces depth expansion with a synergistic design: it propagates long-range context with linear complexity via a grouped state space mixer, denoises and aligns skip features using a cross-scale dual-stage gating bridge, and absorbs cross-site acquisition shifts with a sample-level mixture-of-experts. On the BraTS2019 and BraTS2021 benchmarks, M\textsuperscript{4}Fuse outperforms other lightweight excellent methods in both parameter count and performance. Even at a challenging input resolution of \(64\times128\times128\) (half that of existing excellent models), M\textsuperscript{4}Fuse reduces parameters by 62.63\% and improves average performance by 0.09\%. Ablations of key components validate the method's exceptional parameter-to-accuracy efficiency and robustness across diverse data centers.
\end{abstract}    
\section{Introduction}

Multimodal three dimensional brain tumor segmentation in clinical settings faces three coupled demands \cite{zhu2024brain,an2024dynamic,rasool2025critical}. The model must capture long range three dimensional context in a memory efficient way, it must remain robust under cross site and cross protocol variation, and it must be lightweight for deployment \cite{pan2025vcanet}. The enhancing tumor is contrast dependent and small in volume and thin at the boundary, therefore it is highly sensitive to smoothing from normalization, upsampling, and cross scale fusion, and it is critical for clinical reading \cite{liu2024multimodal,zhou2024m2gcnet}.

In end to end encoder decoder architectures the balance of parameters strongly affects performance and efficiency (a high leverage phenomenon in which small changes in the encoder to decoder ratio can cause large shifts in information flow and optimization). An over heavy encoder with an under powered decoder can produce features that the decoder cannot reconstruct faithfully, which creates a representational bottleneck and wastes encoder capacity \cite{potlapalli2025exploring,liu2025cswin}. The opposite starves early evidence and shifts capacity to later stages, which weakens local cues and can destabilize training. Adding global modules in early encoder layers can increase complexity without commensurate gains \cite{zhou2019high,ronneberger2015u}. Popular plug in components such as efficient decoders and efficient encoders are useful, but blindly reducing complexity on a single side is not a universal solution, because it can harm accuracy, generalization, and multi task or multi expert processing.

\begin{figure}[htbp]
\captionsetup{skip=2pt}
  \centering
  \subfloat[]{  
      \includegraphics[width=0.46\textwidth]{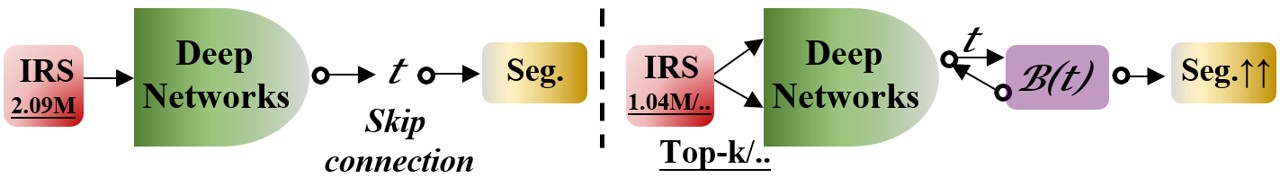}
      \label{fig:Intro2}
  } \\   
  \subfloat[]{  
      \includegraphics[width=0.46\textwidth]{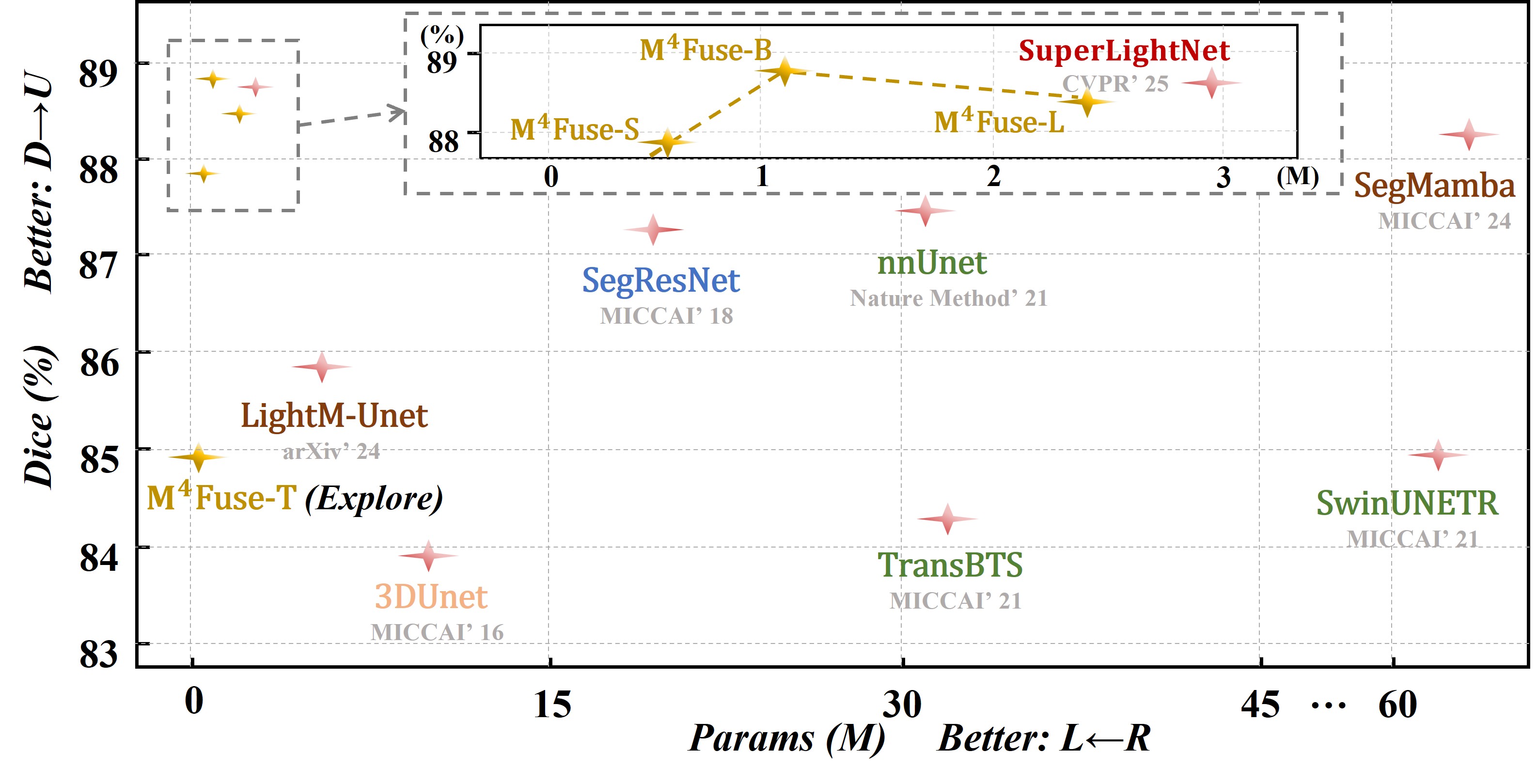}
      \label{fig:Intro1}
  }

  \caption{\small (a) On the left is a standard segmentation architecture, while the right integrates an expert mechanism with a unique attention bridge. The input resolution sequential length (IRS) of 1.04M/2.09M is used to validate the model's generalisation capability. The number of experts is then allocated based on the structures, modalities and types of input data. This is followed by the feature extraction and segmentation processes. The number of allocated experts influences the parameter count and exhibits a positive correlation with it. (b)Visual comparison of lightweight medical segmentation methods versus other excellent methods on the BraTS 2021 dataset, based on Dice metrics, using input resolution sequence lengths as short as 1.04M.}
  \label{fig:Intro}
\end{figure}

We revisit input sizing without assuming a fixed cube. While many recent three dimensional pipelines standardize to \(128\times128\times128\) to accommodate heavy encoders, we also report results at \(128\times128\times128\). We deliberately relax this convention and explore \(64\times128\times128\), which reduces the voxel budget and memory footprint and under our design preserves or improves accuracy. Capacity is placed where it pays off, with a grouped state space mixer supplying long range modeling in linear time while keeping local cues, a cross scale dual stage gating bridge that denoises in space and then realigns channels so the decoder receives clean and consistent skips, and sample level domain experts confined to high semantic low resolution layers, namely at the fourth and fifth encoder stages and at the bottleneck, to absorb center and protocol shifts with minimal overhead. The total number of parameters grows linearly with the number of experts, which keeps scaling controllable.

With purified and aligned skip connections the marginal utility of widening the high semantic or decoder path is low \cite{wang2024narrowing,zhang2024sed}. On the BraTS series our base model with 1.11 million parameters achieves the best or tied Dice and HD95 while wider variants do not deliver significant average gains and can degrade the enhancing tumor. This observation forms a decoder marginal utility principle for medical three dimensional segmentation. Capacity controlled selectivity is more important than widening the decoder.

We propose M\textsuperscript{4}Fuse, a lightweight U shaped framework that embodies these choices. PetaloMixer supplies linear time long range context in mid and late pathways and before each upsampling step. CSBridge applies spatial denoising and cross scale channel realignment so that skip features are consistent and informative. The sample level expert unit adapts only high semantic layers, which avoids dragging early features toward site specific statistics and preserves low complexity at small input resolution.

Our contributions are as follows:
\begin{itemize}
\item We introduce a lightweight and deployment friendly three dimensional framework that combines a grouped state space mixer with a cross scale dual stage gating bridge and achieves strong accuracy with 1.11 million parameters.
\item Sample-level domain experts, confined to high-semantic, low-resolution layers, are introduced, ensuring robustness to site and protocol variation while keeping parameters linearly controllable with expert count.
\item A decoder marginal utility principle for medical three dimensional segmentation is identified, indicating that widening the high semantic or decoder path yields negligible average gains and can harm contrast dependent micro structures such as the enhancing tumor.
\item We present core-parameter ablations and an accuracy-parameter Pareto analysis, showing the base configuration lies on the Pareto frontier in the small-input regime.
\end{itemize}
\section{Related Work}
\label{sec:relatedwork}
\subsection{Multimodal Fusion and Adaptive Routing}
Early multimodal MRI segmentation mainly explored early, middle, and late fusion. Early fusion concatenates modalities at the input and learns a shared encoder \cite{mf4}; middle fusion uses separate encoders and merges features at intermediate or bottleneck layers, preserving modality-specific representations but typically limiting interaction to a single stage and increasing memory and computation \cite{midfusion1,midfusion2}; and late fusion combines deep features or predictions near the output, offering flexibility but weaker access to fine-grained cross-modal cues \cite{latefusion1}. To strengthen interaction, prior work introduced cross-modal attention and gating for spatially adaptive weighting and feature exchange \cite{mf2,mf3,mf1}, dense cross-links for multi-depth feature reuse \cite{mf2}, and cross-modality alignment for unregistered inputs \cite{mf6}. However, these designs often incur high activation and training costs \cite{mf3}, and most model interaction at only one or two scales without explicit cross-scale consistency \cite{mf4,mf1}. Related studies in industrial and vision tasks likewise highlight the value of multiscale guidance, bidirectional cues \cite{mf7}, and application-driven attention \cite{mf5}. MoE routing further decouples modality-specific capacity by selecting experts per sample or region \cite{moe1,moe2,moe3}, reducing cross-modal interference and improving robustness to scanner and site variation \cite{moe6,moe5}; however, expert collapse, load imbalance, and additional latency and memory overhead remain key challenges \cite{rp2,andrews2014overview,zhang2017resilient,moe4,cao2025moe}. Related task-adaptive routing has also shown promise in medical image restoration \cite{rp1}, but its integration with multi-scale fusion and decoder skip connections remains underexplored.

\subsection{Lightweight Multimodal Brain Segmentation}
Transformer token mixers such as UNETR and Swin UNETR improve volumetric segmentation by modeling long-range context, but remain expensive in 3D due to the cost of attention over many tokens and the need for deep stacks \cite{swinunetv2,ronneberger2015u,TransBTS}. Efficient attention reduces complexity through factorization or linearization, yet stable, budget-matched high-resolution performance is still difficult \cite{xu2023lightweight,lightunet,zhou2025dcl}. State space models provide linear-complexity sequence mixing and have been adopted in 3D encoders and U-shaped backbones \cite{SegMamba}, but key issues remain in preserving 3D locality, aggregating multiple scan directions, and integrating sequence mixing into decoder stages dominated by upsampling and multi-scale fusion. In parallel, lightweight 3D CNN designs and compression techniques reduce parameters and FLOPs through efficient convolutions, pruning, and distillation \cite{SuperLightNet,damnet,3dunet,liu2021paddleseg}, but may weaken channel interaction, global reasoning, and robustness to domain shift, especially for small lesions and boundaries \cite{cao2025lcmf}. Accordingly, BraTS evaluation still centers on Dice and HD95 \cite{nnUnet}, while efficiency metrics such as parameters, FLOPs, and latency are becoming increasingly important in lightweight segmentation.

\section{Method}
\label{sec:method}

\begin{figure*}[htbp]
    \centering
    \includegraphics[width=\textwidth]{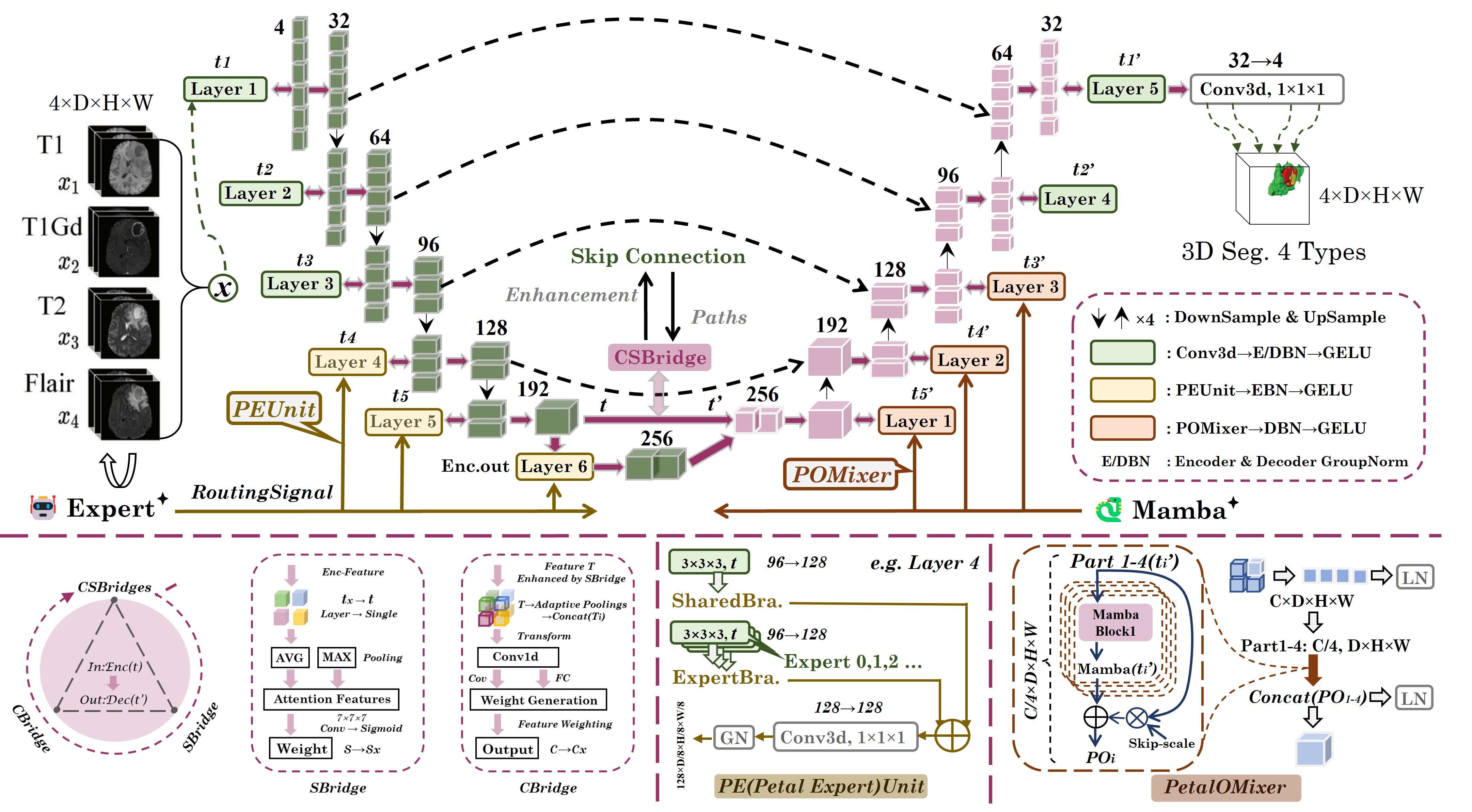}
    \caption{{\small \textbf{Overall architecture of M\textsuperscript{4}Fuse.} 
    (a) CSBridge (CSB, feature fusion stage): CSB connects the encoder and decoder to enhance multi-scale features by integrating the spatial-channel attention of SBridge and CBridge:
    $\mathrm{Dec}(t')=\mathrm{Enc}(t)+t\cdot\mathrm{SBridge}(t)+\mathrm{CBridge}(t)$,
    where $t' = t \cdot s_x + c_x + t_x$ is fed to the decoder. Each decoder stage fuses with its corresponding subsampled encoder feature through residual skip connections. Here, skip connections provide propagation paths, while CSB acts as a feature enhancer to mitigate detail loss from encoder downsampling and insufficient reference cues during decoder upsampling. 
    (b) Petal Expert Unit (PEU): Located at the encoder backend, PEU uses a shared-branch plus expert-branch design to handle multimodal and multidataset variation, enabling the encoder to capture both general anatomy and dataset-specific lesion patterns. 
    (c) POMixer (POM): Placed at the decoder frontend, POM treats 3D features as sequences and uses Mamba to model long-range dependencies while jointly aggregating spatial and channel information, helping the decoder preserve local details and global semantics during high-resolution reconstruction.}}
    \label{fig:main}
\end{figure*}

\subsection{Overview}
Volumetric brain tumor segmentation with multimodal MRI requires three properties at once. The network must propagate long-range context across depth, height, and width in a memory-efficient way. The decoder must receive skip connections that are denoised and consistent across scales to avoid redundant or conflicting activations. The representation must adapt across acquisition differences while keeping the parameter budget very small (see Figure~\ref{fig:main}). We introduce M\textsuperscript{4}Fuse, which realizes these properties with a lightweight U-shaped backbone. A grouped state-space mixer (POM) provides global context in the mid and late pathway. A cross-scale dual-stage gating bridge (CSB) purifies and aligns encoder features before decoding. A shared-plus-expert unit (PEU) adapts high-level features across centers.
The input is \(x\in\mathbb{R}^{B\times C\times D\times H\times W}\) where \(B\) is the mini-batch size, \(C\) is the number of input channels corresponding to modalities, \(D,H,W\) are spatial sizes, and \(C=4\) on BraTS. Encoder outputs at five scales are \(t_s\in\mathbb{R}^{B\times C_s\times D_s\times H_s\times W_s}\) for \(s\in\{1,2,3,4,5\}\) with resolution decreasing as \(s\) increases. The deepest representation is the bottleneck \(b\). The bridge transforms \(\{t_s\}_{s=1}^{5}\) into \(\{\tilde t_s\}_{s=1}^{5}\) which are reused in the decoder. The decoder proceeds from \(b\) through three stages \(d_1,d_2,d_3\). We denote decoder activations by \(d_k\in\mathbb{R}^{B\times C_{d_k}\times D_{d_k}\times H_{d_k}\times W_{d_k}}\) for \(k\in\{1,2,3\}\). At each stage it first applies the mixer, then upsamples, then fuses with the bridged feature at the matched resolution, so \(d_1\) fuses with \(\tilde t_5\), \(d_2\) fuses with \(\tilde t_4\), \(d_3\) fuses with \(\tilde t_3\). The expert unit, applied at \(t_4\), \(t_5\), and \(b\) and its runtime scales linearly with the number of voxels.

\subsection{Grouped State Space Mixer}
A three-dimensional tensor is reshaped into a sequence that preserves voxel order in a fixed raster-scan order. Layer normalization is \(\mathrm{LN}(\cdot)\) and LN acts on the channel axis. The reshape stacks spatial locations into length \(L\). The transpose \((\cdot)^\top\) swaps the last two axes so that the sequence length becomes the second axis. Concatenation along channels is \([\cdot\,\|\,\cdot]\). We write
\[
\left\{
\begin{aligned}
X &= \mathrm{LN}\!\big(\mathrm{reshape}(x; B,C,L)\big)^\top \in \mathbb{R}^{B\times L\times C},\\
L &= D\,H\,W,
\end{aligned}
\right.
\tag{1}
\]
where \(L\) is the number of voxels and \(C\) is the channel width. The channel axis is partitioned into \(g\) equal groups and we use \(g=4\). Let
\[
X=[X^{(1)} \,\|\, \dots \,\|\, X^{(g)}],\quad X^{(j)}\in\mathbb{R}^{B\times L\times C/g}.
\]
Each group is scanned by a discrete state-space model along the sequence index \(k\in\{1,\dots,L\}\). The hidden state is \(h^{(j)}_{k}\in\mathbb{R}^{B\times d^{(j)}}\), where \(d^{(j)}\) is the state dimension for group \(j\). The system matrices are \(A^{(j)}\in\mathbb{R}^{d^{(j)}\times d^{(j)}}\), \(B^{(j)}\in\mathbb{R}^{d^{(j)}\times (C/g)}\), \(C^{(j)}\in\mathbb{R}^{(C/g)\times d^{(j)}}\). With step size \(\Delta>0\) the discrete recursion is
\[
\left\{
\begin{aligned}
h^{(j)}_{k+1} &= \bar A^{(j)} h^{(j)}_{k} + \bar B^{(j)} X^{(j)}_{k},\\
Y^{(j)}_{k} &= C^{(j)} h^{(j)}_{k},
\end{aligned}
\right.
\qquad k=1,\dots,L,
\tag{2}
\]
where \(Y^{(j)}_{k}\in\mathbb{R}^{B\times (C/g)}\) is the group output at position \(k\). The matrices \(\bar A^{(j)}\in\mathbb{R}^{d^{(j)}\times d^{(j)}}\) and \(\bar B^{(j)}\in\mathbb{R}^{d^{(j)}\times (C/g)}\) are the discretized transition and input maps obtained from a continuous linear system by a matrix exponential. The matrix exponential is \(\exp(\cdot)\). We write
\[
\left\{
\begin{aligned}
\bar A^{(j)} &= \exp\!\big(A^{(j)}\Delta\big),\\
\bar B^{(j)} &= \displaystyle\int_{0}^{\Delta}\exp\!\big(A^{(j)}\tau\big)\,B^{(j)}\,d\tau.
\end{aligned}
\right.
\tag{3}
\]
Stacking over \(k\) yields \(\mathrm{SSM}(X^{(j)})\in\mathbb{R}^{B\times L\times (C/g)}\) whose \(k\)th slice equals \(Y^{(j)}_{k}\). A learnable nonnegative residual scale \(s\in\mathbb{R}_{\ge 0}\) stabilizes narrow widths
\[
Z^{(j)}=\mathrm{SSM}\!\big(X^{(j)}\big)+s\,X^{(j)}.
\tag{4}
\]
Group outputs are concatenated, normalized, and linearly projected with \(W\in\mathbb{R}^{C\times C_{\mathrm{out}}}\) where \(C_{\mathrm{out}}\) is the output width
\[
\left\{
\begin{aligned}
Z &= \mathrm{LN}\!\big([Z^{(1)} \,\|\, \dots \,\|\, Z^{(g)}]\big)\in\mathbb{R}^{B\times L\times C},\\
U &= Z\,W\in\mathbb{R}^{B\times L\times C_{\mathrm{out}}}.
\end{aligned}
\right.
\tag{5}
\]
Finally \(U\) is reshaped to \(\mathbb{R}^{B\times C_{\mathrm{out}}\times D\times H\times W}\).
The scan in each group has cost \(\mathcal{O}(L\,C/g)\) so the mixer is linear in \(L\). Attention has cost \(\mathcal{O}(L^2 C)\). The mixer is inserted at \(t_4\), \(t_5\), and \(b\) and at each decoder stage before upsampling to keep global information along the reconstruction path.
Let \(\rho(\cdot)\) be the spectral radius and let \(\|\cdot\|\) be any submultiplicative operator norm. Let \(\|X^{(j)}\|:=\sup_{k}\|X^{(j)}_{k}\|\). If \(\rho(\bar A^{(j)})\le 1\) then, using submultiplicativity and the uniform bound on \(\|X^{(j)}_k\|\), the sequence defined by \((2)\) and \((4)\) satisfies
\[
\left\{
\begin{aligned}
\|Z^{(j)}\| &\le \kappa\,\|X^{(j)}\| + s\,\|X^{(j)}\|,\\
\kappa &= \sup_{k}\ \|C^{(j)}\|\ \sum_{t=0}^{k}\ \|\bar A^{(j)}\|^{t}\ \|\bar B^{(j)}\|.
\end{aligned}
\right.
\tag{6}
\]
When \(\|\bar A^{(j)}\|<1\) the sum is dominated by a geometric bound. The direct term with scale \(s\) prevents vanishing or exploding responses in long scans and the normalizations in \((1)\) and \((5)\) bound feature-scale drift.

\subsection{Cross Scale Dual Stage Gating Bridge}
Skip connections pass fine structure yet they also propagate modality-specific noise and scanner artifacts. We purify and align encoder features before decoding in two stages. The sigmoid is \(\sigma(\cdot)\). Element-wise multiplication is \(\odot\). The operator \(\mathrm{Conv}_{7}\) is a three-dimensional convolution with kernel size seven that maps two input channels to one attention map. The operators \(\mathrm{Avg}(t_s)\) and \(\mathrm{Max}(t_s)\) are channel-wise average and maximum giving tensors in \(\mathbb{R}^{B\times 1\times D_s\times H_s\times W_s}\). For each scale \(s\) a spatial attention mask is computed and applied
\[
\left\{
\begin{aligned}
a_s &= \sigma\!\Big(\mathrm{Conv}_{7}\big([\mathrm{Avg}(t_s),\,\mathrm{Max}(t_s)]\big)\Big)\in\mathbb{R}^{B\times 1\times D_s\times H_s\times W_s},\\
t^{\mathrm{sp}}_s &= a_s\odot t_s\in\mathbb{R}^{B\times C_s\times D_s\times H_s\times W_s}.
\end{aligned}
\right.
\tag{7}
\]
Global average pooling over spatial axes is \(\mathrm{GAP}(\cdot)\). Per-scale channel statistics are \(z_s=\mathrm{GAP}(t^{\mathrm{sp}}_s)\in\mathbb{R}^{B\times C_s}\). Concatenation over scales gives \(z=[z_1 \,\|\, z_2 \,\|\, z_3 \,\|\, z_4 \,\|\, z_5]\in\mathbb{R}^{B\times C_{\Sigma}}\) where \(C_{\Sigma}=\sum_{s}C_s\). For each scale a channel gate uses weights \(W_s\in\mathbb{R}^{C_{\Sigma}\times C_s}\) and bias \(b_s\in\mathbb{R}^{C_s}\)
\[
g_s=\sigma\!\big(z\,W_s+b_s\big)\in\mathbb{R}^{B\times C_s}.
\tag{8}
\]
The vector \(g_s\) is broadcast along spatial axes to match \(t^{\mathrm{sp}}_s\). Two nonnegative learned scalars \(\alpha\) and \(\beta\) control residual strength
\[
\left\{
\begin{aligned}
\hat t_s &= t_s+\alpha\,t^{\mathrm{sp}}_s+\beta\,(g_s\odot t^{\mathrm{sp}}_s),\\
\alpha &\ge 0,\quad \beta \ge 0.
\end{aligned}
\right.
\tag{9}
\]
\noindent\textit{Since} \(\sigma:\mathbb{R}\!\to\!(0,1)\), \textit{we have} \(\|a_s\|_{\infty}\le 1\) \textit{and} \(\|g_s\|_{\infty}\le 1\).
Using any submultiplicative norm, the bridge is bounded as
\[
\|\hat t_s\|\le (1+\alpha+\beta)\,\|t_s\|.
\tag{10}
\]
The bridge is applied once to \(\{t_1,t_2,t_3,t_4,t_5\}\) and the bridged tensors are reused in the decoder.

\begin{algorithm}[t]
\caption{M\textsuperscript{4}Fuse Forward Pass}
\label{alg:m4fuse_forward}
\begin{algorithmic}[1]
\State \textbf{Input} volume \(x\in\mathbb{R}^{B\times C\times D\times H\times W}\): \(B\) mini-batch size, \(C\) modalities, \(D,H,W\) spatial sizes
\State \textbf{Input} id vector \(d\in\{1,\dots,M\}^{B}\): one expert index per sample
\State \textbf{Input} operators: \(\mathrm{Conv3D}\) (3D conv), \(\mathrm{GN}\) (group norm), \(\mathrm{Pool}\) (strided pooling), \(\mathrm{Up}\) (trilinear upsampling), \(\mathrm{POM}\) (grouped state-space mixer), \(\mathrm{PEU}\) (shared-plus-expert unit), \(\mathrm{CSB}\) (cross-scale dual-stage bridge), \(\mathrm{Conv1\times1\times1}\) (pointwise 3D conv)
\State \(t_1=\mathrm{GN}(\mathrm{Conv3D}(x))\)
\State \(t_2=\mathrm{GN}(\mathrm{Conv3D}(\mathrm{Pool}(t_1)))\)
\State \(t_3=\mathrm{GN}(\mathrm{Conv3D}(\mathrm{Pool}(t_2)))\)
\State \(u_4=\mathrm{Pool}(t_3)\)
\State \(t_4=\mathrm{GN}\!\left(\mathrm{PEU}_4(u_4, d)\right)\)
\State \(u_5=\mathrm{Pool}(t_4)\)
\State \(t_5=\mathrm{GN}\!\left(\mathrm{PEU}_5(u_5, d)\right)\)
\State \(b=\mathrm{PEU}_b\!\left(\mathrm{Pool}(t_5), d\right)\)
\State \((\tilde t_1,\tilde t_2,\tilde t_3,\tilde t_4,\tilde t_5)=\mathrm{CSB}(t_1,t_2,t_3,t_4,t_5)\)
\State \(y_1=\mathrm{GN}(\mathrm{POM}(b))+\tilde t_5\)
\State \(y_1=\mathrm{Up}(y_1)\)
\State \(y_2=\mathrm{GN}(\mathrm{POM}(y_1))+\tilde t_4\)
\State \(y_2=\mathrm{Up}(y_2)\)
\State \(y_3=\mathrm{GN}(\mathrm{POM}(y_2))+\tilde t_3\)
\State \(y_3=\mathrm{Up}(y_3)\)
\State \(y_4=\mathrm{GN}(\mathrm{Conv3D}(y_3))+\tilde t_2\)
\State \(y_4=\mathrm{Up}(y_4)\)
\State \(y_5=\mathrm{GN}(\mathrm{Conv3D}(y_4))+\tilde t_1\)
\State \textbf{Output} logits \(\hat y=\mathrm{Conv1\times1\times1}(y_5)\)
\end{algorithmic}
\end{algorithm}

\subsection{Shared Plus Expert Routing}
Acquisition protocols differ across centers and cohorts. A single shared pathway is often suboptimal whereas a fully specialized network is heavy. We adopt a compact shared-plus-expert unit at high semantic layers. The shared mixer is \(f_{\mathrm{sh}}\). The expert mixers are \(\{f_m\}_{m=1}^{M}\). For sample \(i\) the high-level input is \(u_i\in\{t_4,t_5,b\}\). A top-one expert index is \(\pi(u_i)\in\{1,\dots,M\}\) which can be assigned by a dataset identifier. The unit output for sample \(i\) is
\[
\mathrm{PEU}(u_i)=f_{\mathrm{sh}}(u_i)+ f_{\pi(u_i)}(u_i).
\tag{11}
\]
A top-\(K\) variant averages the selected experts’ outputs. To stabilize gradients, we use batch-wise concatenation instead of in-place assignment, with dropout \(p\in[0,1)\)
\[
\mathrm{PEU}(u)=\mathrm{Dropout}_p\Big(f_{\mathrm{sh}}(u)+\mathrm{concat}_{i}\big(f_{\pi(u_i)}(u_i)\big)\Big).
\tag{12}
\]
Here \(\mathrm{concat}_{i}\) denotes concatenation along the batch dimension. Algorithm~\ref{alg:m4fuse_forward} summarizes the forward path and Algorithm~\ref{alg:fullbloom} summarizes the bridge.

\begin{algorithm}[t]
\caption{Cross-scale dual-stage gating bridge}
\label{alg:fullbloom}
\begin{algorithmic}[1]
\State \textbf{Input} features \((t_1,t_2,t_3,t_4,t_5)\) with shapes \(t_s\in\mathbb{R}^{B\times C_s\times D_s\times H_s\times W_s}\)
\State \textbf{Input} weights \(\{W_s\}\) with \(W_s\in\mathbb{R}^{C_{\Sigma}\times C_s}\); biases \(\{b_s\}\) with \(b_s\in\mathbb{R}^{C_s}\); scalars \(\alpha,\beta\) with \(\alpha\ge 0,\ \beta\ge 0\)
\State \textbf{Input} operators: \(\sigma\) (sigmoid), \(\odot\) (element-wise product), \(\mathrm{Conv}_7\) (three-dimensional conv, kernel size seven), \(\mathrm{GAP}\) (global average pooling), \(\mathrm{Avg}\) (channel-wise average), \(\mathrm{Max}\) (channel-wise maximum), \(\mathrm{concat}\) (channel-wise concatenation)
\For{\(s\in\{1,2,3,4,5\}\)}
  \State \(a_s=\sigma\big(\mathrm{Conv}_7([\mathrm{Avg}(t_s),\mathrm{Max}(t_s)])\big)\)
  \State \(t^{\mathrm{sp}}_s=a_s\odot t_s\)
\EndFor
\State \(z=\mathrm{concat}(\mathrm{GAP}(t^{\mathrm{sp}}_1),\dots,\mathrm{GAP}(t^{\mathrm{sp}}_5))\)
\For{\(s\in\{1,2,3,4,5\}\)}
  \State \(g_s=\sigma(z W_s+b_s)\);\quad broadcast \(g_s\) to shape of \(t^{\mathrm{sp}}_s\)
  \State \(\hat t_s=t_s+\alpha\,t^{\mathrm{sp}}_s+\beta\,(g_s\odot t^{\mathrm{sp}}_s)\)
\EndFor
\State \textbf{Output} \((\hat t_1,\hat t_2,\hat t_3,\hat t_4,\hat t_5)\)
\end{algorithmic}
\end{algorithm}

\subsection{Network Variants}

{M\textsuperscript{4}Fuse (Input Resolution: 128 → 4) has three variants:

\begin{itemize}[leftmargin=2em, labelsep=0.5em, topsep=5pt, itemsep=4pt]
    \item \textit{M\textsuperscript{4}Fuse-T}: \( Max(C)=128 \), \textit{SEL} \(= \{128, 4, 4, 4\} \)
    \item \textit{M\textsuperscript{4}Fuse-S}: \( Max(C)=196 \), \textit{SEL} \(= \{196, 4, 4, 4\} \)
    \item \textit{M\textsuperscript{4}Fuse-B}: \( Max(C)=256 \), \textit{SEL} \(= \{256, 4, 4, 4\} \)
    \item \textit{M\textsuperscript{4}Fuse-L}: \( Max(C)=384 \), \textit{SEL} \(= \{384, 4, 4, 4\} \)
\end{itemize}
where C and SEL represent the maximum number of network channels and the Size of the Enc.out Layer (SEL), respectively.

\begin{figure*}[htbp]
    \centering
    \includegraphics[width=\textwidth]{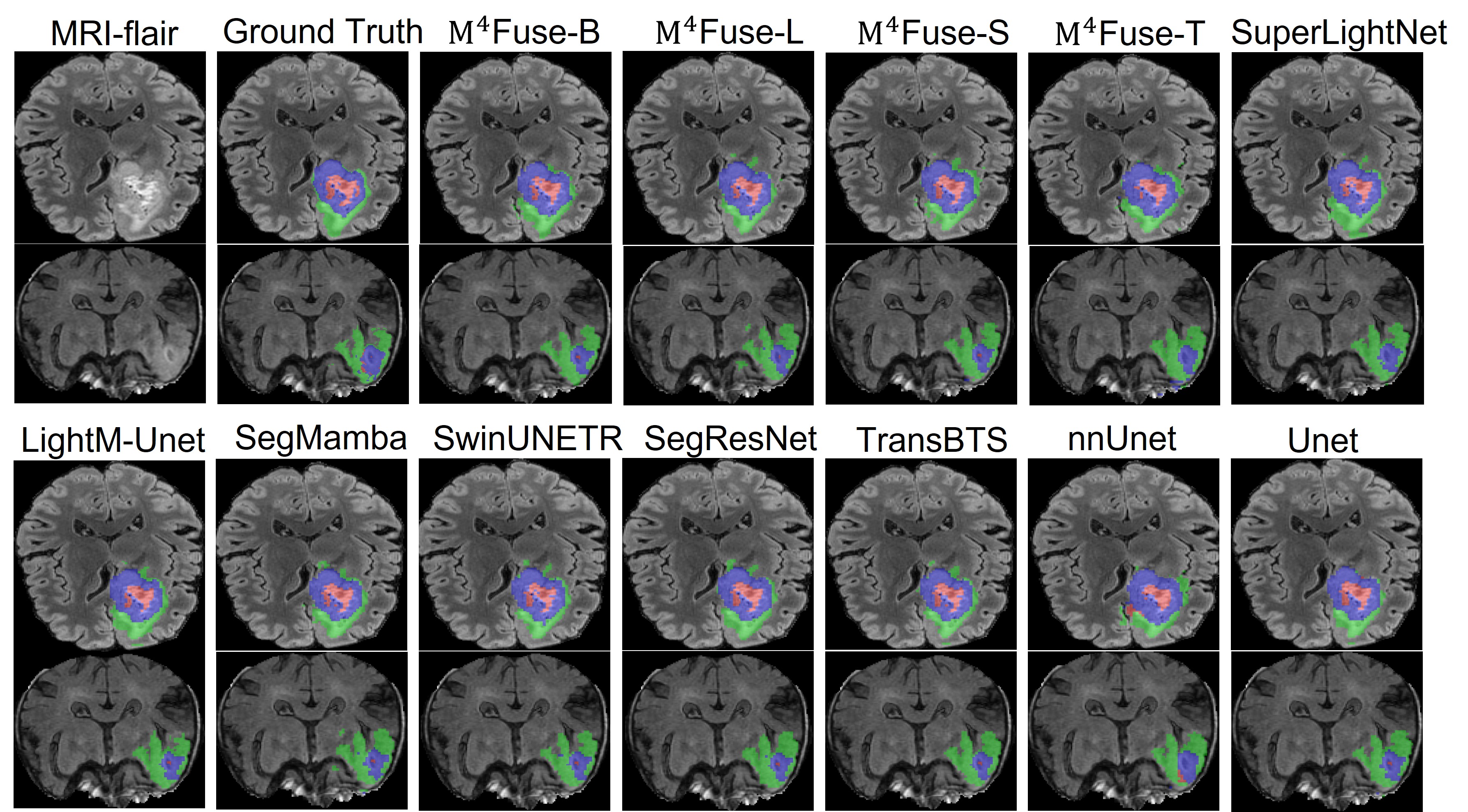}
    \caption{{\small Visualized segmentation of BraTS 2021 datasets, input resolution: 64×128×128, where red indicates tumor core, blue indicates whole tumor, and green indicates enhancing tumor.}}
    \label{fig:VisCo}
\end{figure*}

\section{Experiments}

\textbf{Datasets.} The proposed M\textsuperscript{4}Fuse model was trained and tested on two challenging multimodal brain tumor datasets (BraTS2019, BraTS2021), with each 3D brain MRI volume containing four modalities (T1, T1ce, T2, T2-FLAIR) and three segmentation targets (WT: Whole Tumor, ET: Enhancing Tumor, TC: Tumor Core). BraTS2019 (335 volumes, low/high-grade [LGG/HGG] gliomas) serves as the core reference for comparisons, quantifications, and ablation benchmarks. BraTS2021 adopts unified data grading to extend low-resolution scenarios, with its visual comparisons further validating the model’s lightweight, efficiency, and generalization. Given inter-dataset grading discrepancies, different expert counts were configured during training/comparisons (Note: The expert mechanism adapts to lesion type, location, and modality variations across datasets; see \textbf{Supplementary Table S2.1}).

BraTS2019 used five-fold cross-validation (80\% training, 20\% validation per fold) to ensure result stability and reliability. BraTS2021 data was randomly split into training/validation/test sets at a 6:2:2 ratio, with the validation set excluded from gradient training.

\begin{table*}[htbp]
    \centering
    
    \setlength{\tabcolsep}{3.0pt}
    \renewcommand{\arraystretch}{1.1} 
    \captionsetup{aboveskip=5pt}
    \caption{Performance comparison with SOTA methods on BraTS2019 validation set using 5-Cross-Validation Mode and BraTS2021 validation set using Train-Valid-Eval Mode. The mean performance (Dice \%, HD95 mm and \#params M) of all models combinations is shown here (due to space limit, the detailed each stage from different modes performance of 5-fold and TvE metrics is given in \textbf{supplementary Table S1}). Top-k=1/2 signifies the MoE module's specialized extraction and sharing of training expertise across data types or modalities, while simultaneously enabling internal comparisons between datasets.} 
    \label{tab:com1}
    \resizebox{1.0\textwidth}{!}{  
        
        \begin{tabular}{l c |cccc cccc| cccc cccc} 
            & & \multicolumn{8}{c|}{BraTS2021 (Top-k=1, IRS=1.04M)}  & \multicolumn{8}{c}{BraTS2019 (Top-k=2, IRS=2.09M)} \\ 
            \cmidrule[0.5pt]{3-10}
            \cmidrule[0.5pt]{11-18}  
            
            \multirow{1}{*}{Methods} & \multirow{1}{*}{\shortstack{\textit{Params}\\ \textit{(M)}}} & \multicolumn{4}{c}{\textit{Dice}↑\ \textit{(\%)}} & \multicolumn{4}{c|}{\textit{HD95}↓\ \textit{(mm)}} & \multicolumn{4}{c}{\textit{Dice}↑\ \textit{(\%)}} & \multicolumn{4}{c}{\textit{HD95}↓\ \textit{(mm)}} \\

            & & WT & TC & ET & Avg & WT & TC & ET & Avg & WT & TC & ET & Avg & WT & TC & ET & Avg \\ 
            \cmidrule[0.5pt]{1-18}
            3D Unet \cite{3dunet} & 12.34 & 87.91 & 85.61 & 78.22 & \cellcolor{gray!10}83.91 & 3.81 & 3.32 & 3.11 & \cellcolor{gray!10}3.41 & 84.72 & 74.07 & 67.49 & \cellcolor{gray!10}75.42 & 8.81 & 13.58 & 13.43 & \cellcolor{gray!10}9.48 \\ 
            
            nnUnet \cite{nnUnet} & 31.19 & 89.26 & 87.58 & 84.77 & \cellcolor{gray!10}87.34 & 3.20 & 2.55 & 2.89 & \cellcolor{gray!10}2.88 & 87.43 & 79.71 & 76.79 & \cellcolor{gray!10}81.31 & 4.87 & 5.16 & 5.29 & \cellcolor{gray!10}5.10 \\ 
            
            TransBTS \cite{TransBTS} & 31.65 & 88.43 & 85.78 & 78.58 & \cellcolor{gray!10}84.26 & 6.63 & 3.61 & 3.03 & \cellcolor{gray!10}4.42 & 87.52 & 79.23 & 76.22 & \cellcolor{gray!10}80.99 &  4.68 &  4.98 &  3.99 & \cellcolor{gray!10}4.55 \\ 

            SegResNet \cite{SegResNet} & 18.80 & 89.59 & 89.42 & 82.67 & \cellcolor{gray!10}87.22 & 3.17 & 2.64 & 2.47 & \cellcolor{gray!10}2.76 & 88.71 & 82.08 & 74.28 & \cellcolor{gray!10}81.69 & 3.99 & 5.48 & 5.02 & \cellcolor{gray!10}4.83 \\ 

            SwinUNETR \cite{swinunetv2} & 62.19 & 89.29 & 86.74 & 78.88 & \cellcolor{gray!10}84.97 & 3.24 & 2.65 & 2.74 & \cellcolor{gray!10}2.87 & 88.16 & 80.73 & 75.82 & \cellcolor{gray!10}81.57 & 4.00 & 4.93 & 5.17 & \cellcolor{gray!10}4.70 \\ 
            
            SegMamba \cite{SegMamba} & 66.85 & 90.48 & 85.54 & 84.01 & \cellcolor{gray!10}88.23 & 3.97 & 2.55 & 2.12 & \cellcolor{gray!10}2.88 & 89.11 & 82.25 & 75.71 & \cellcolor{gray!10}82.35 & 4.28 & 5.65 & 5.77 & \cellcolor{gray!10}5.23 \\ 
            
            LightM-Unet \cite{lightunet} & 5.02 & 89.99 & 87.27 & 80.45 & \cellcolor{gray!10}85.90 & 3.47 & 2.48 & 2.34 & \cellcolor{gray!10}2.76 & 88.70 & 80.20 & 72.43 & \cellcolor{gray!10}80.44 & 4.68 & 5.18 & 4.67 & \cellcolor{gray!10}4.84 \\ 
            
            SuperLightUnet \cite{SuperLightNet} & 2.97 & 90.57 & 89.23 & 86.33 & \cellcolor{gray!10}88.70 & 3.40 & 2.53 & 2.11 & \cellcolor{gray!10}2.68 & 88.54 & 80.64 & 74.32 & \cellcolor{gray!10}81.16 & 4.74 & 6.36 & 5.33 & \cellcolor{gray!10}5.47 \\ \cmidrule[0.5pt]{1-18}
        
            \multirow{4}{*}{\rotatebox{90}{Top-k=1/2}} 
            \ M\textsuperscript{4}Fuse-T & 0.29/0.32 & 87.01 & 85.97 & 81.73 & \cellcolor{gray!10}84.90 & 4.08 & 3.18 & 2.56 & \cellcolor{gray!10}3.27 & 88.31 & 81.00 & 72.95 & \cellcolor{gray!10}80.75 & 3.70 & 4.88 & 4.07 & \cellcolor{gray!10}4.21 \\ 
            
            \ \ \ \ \ \ M\textsuperscript{4}Fuse-S & 0.63/0.70 & 88.82 & 88.23 & 86.72 & \cellcolor{gray!10}87.92 & 4.25 & 2.70 & 2.12 & \cellcolor{gray!10}3.02 & 88.66 & 82.12 & 74.38 & \cellcolor{gray!10}81.72 & 5.41 & 5.52 & 4.93 & \cellcolor{gray!10}5.28 \\ 
            
            \ \ \ \ \ \ M\textsuperscript{4}Fuse-B & \cellcolor{gray!20}\textbf{1.11/1.23} & \cellcolor{gray!20}\textbf{89.14} & \cellcolor{gray!20}\textbf{89.74} & \cellcolor{gray!20}\textbf{87.49} & \cellcolor{gray!20}\textbf{88.79} & \cellcolor{gray!20}\textbf{3.66} & \cellcolor{gray!20}\textbf{2.42} & \cellcolor{gray!20}\textbf{1.90} & \cellcolor{gray!20}\textbf{2.66} & \cellcolor{gray!20}\textbf{89.31} & \cellcolor{gray!20}\textbf{82.69} & \cellcolor{gray!20}\textbf{75.16} & \cellcolor{gray!20}\textbf{82.38} & \cellcolor{gray!20}\textbf{4.14} & \cellcolor{gray!20}\textbf{4.46} & \cellcolor{gray!20}\textbf{4.75} & \cellcolor{gray!20}\textbf{4.51} \\ 
            
            \ \ \ \ \ \ M\textsuperscript{4}Fuse-L & 2.45/2.72 & 89.33 & 90.17 & 85.89 & \cellcolor{gray!10}88.46 & 3.83 & 2.57 & 1.88 & \cellcolor{gray!10}2.76 & 88.89 & 81.74 & 75.51 & \cellcolor{gray!10}82.04 & 3.30 & 4.50 & 4.30 & \cellcolor{gray!10}4.03 \\ 
            
        \end{tabular}
        }
\end{table*}


\textbf{Implementation details.} Our model was implemented in Python 3.8, PyTorch 2.0, and MONAI 1.3, with training and inference conducted on A100 GPUs. For the training and validation of M\textsuperscript{4}Fuse’s, the BraTS 2019 and BraTS 2021 datasets were randomly cropped to fixed sizes of 128×128×128 (IRS=2.09M) and 64×128×128 (IRS=1.04M) respectively, with a unified batch size of 2 adopted for both datasets. The AdamW optimizer was employed with an initial learning rate of 1e-4, a weight decay of 1e-5 and a minimum learning rate of 1e-6, whose learning rate was dynamically adjusted by a cosine annealing scheduler during training. The loss function was constructed as a weighted combination of Dice Loss and Cross Entropy Loss with a weight ratio of 7:3. For dataset partitioning, 5-Fold Cross-Validation was used for BraTS 2019, and the BraTS 2021 dataset was randomly split into training, validation and test sets at a ratio of 6:2:2; early stopping with a patience of 80 (monitoring total Dice coefficient) was applied for the training of BraTS 2019 to avoid overfitting. The total training epochs were set to 200 for BraTS 2019 and 300 for BraTS 2021, with gradient scaling (AMP) and TF32 acceleration techniques utilized to optimize training efficiency. In the inference phase, Test Time Augmentation was adopted to enhance the robustness of the model's prediction results.

\subsection{Comparison with others}
Table~\ref{tab:com1} shows experimental results on BraTS2019 and BraTS2021.M\textsuperscript{4}Fuse achieves excellent segmentation performance and parameter efficiency, with Dice scores of 82.38\% and 88.79\%,  surpassing SegMamba (82.35\%) and SuperLightUnet (88.70\%) by 0.03\% and 0.09\%, respectively. With 62.63\% fewer parameters than SuperLightUnet, it significantly outperforms other models. Additionally, this model achieves the best HD95 metric on the BraTS2019 dataset. 

Figure~\ref{fig:VisCo} shows qualitative segmentation comparisons on BraTS2021 (Flair modality as background, two samples). M\textsuperscript{4}Fuse converges more stably and rapidly during training. The upper panel reveals lower ET region tissue segmentation loss and more precise segmentation and the lower panel demonstrates superior WT/ET region consistency, boundary handling, and reduced oversegmentation.

To address the core parameter count issue, we quantified and compared structural parameter distributions across models from Table~\ref{tab:com2}. Notably, only 3D Unet, nnUnet, SwinUNETR, and our model show balanced parameter allocations across modules, validating architectural robustness.  Most lightweight models rely on plug-and-play modules (e.g., Mobile backbones for encoders, FLD \cite{liu2021paddleseg} for decoders), merely assembling components to achieve the final overall module quantification—failing to achieve fundamental lightweighting. In contrast, our method embeds diverse mechanisms in UNet, such as data-adaptive specialized modules and the CSBridge attention enhancer (integrated with skip connections to maximize encoder feature utilization for segmentation). Finally, our Base model shows consistent segmentation metric decline with reduced input resolution. For Training/Inference Memory and GFLOPS comparisons, see \textbf{Supplementary Table S3}.

\begin{table}[htbp]
    \centering
    \small
    \setlength{\tabcolsep}{-1.5pt} 
    \captionsetup{aboveskip=5pt}
    \caption{\small Quantitative evaluations with 4×128\textsuperscript{3} (IRS=2.09M).} 
    \label{tab:com2}
    \renewcommand{\arraystretch}{1.0} 
    \hspace{-0.5em}
    \resizebox{0.48\textwidth}{!}{
    \begin{tabular}{l|ccc|cc}
        \multirow{2}{*}{Methods} &
        \multicolumn{3}{c|}{\makecell{\textit{Parameter Structure (M)}}} & 
        \multirow{2}{*}{\textit{In\ Res}} &  
        \multirow{2}{*}{\textit{Params}} \\  

        & \textit{\#encoder} & \textit{\#decoder} & \textit{else (CSB/.)}   & & \\  
        \cmidrule[0.5pt]{1-6}  
            3D Unet          & \underline{6.12 (49.59\%)} & \underline{6.22 (50.41\%)} & - & \textit{128\textsuperscript{3}} & 12.34   \\ 
            nnUnet         & \underline{5.09 (16.33\%)} & \underline{3.07 (9.85\%)} & \underline{23.02 (73.81\%)}  & \textit{128\textsuperscript{3}}& 31.19   \\ 
            TransBTS       & 31.58 (99.81\%) & 0.07 (0.19\%) & - & \textit{128\textsuperscript{3}}& 31.65   \\ 

            SegResNet       & 17.59 (93.56\%) & 1.12 (6.44\%) & - & \textit{128\textsuperscript{3}}& 18.80    \\ 

            SwinUNETR       & \underline{13.70 (22.03\%)} & \underline{16.94 (27.24\%)} & \underline{31.54 (50.73\%)} & \textit{128\textsuperscript{3}}& 62.19    \\ 

            SegMamba       & 60.87 (91.06\%) & 5.97 (8.94\%) & - & \textit{128\textsuperscript{3}}& 66.85    \\ 
            
            LightM-Unet     & 4.94 (98.53\%) & 0.08 (1.47\%) & - & \textit{128\textsuperscript{3}}& 5.02    \\ 
            SuperLightUnet  & 2.70 (91.07\%)& 0.26 (8.88\%) & - & \textit{128\textsuperscript{3}}& 2.97   \\ 
            \cmidrule[0.5pt]{1-6}

            \cellcolor{gray!20}M\textsuperscript{4}Fuse-T \textit{(0.5×)}   & \cellcolor{gray!20}0.12 (43.3\%) & \cellcolor{gray!20}0.09 (33.6\%) & \cellcolor{gray!20}0.06 (23.0\%) & \cellcolor{gray!20}\textit{128\textsuperscript{3}}&  \cellcolor{gray!20}0.32    \\ 

            \cellcolor{gray!20}M\textsuperscript{4}Fuse-S \textit{(0.75×)}  & \cellcolor{gray!20}0.27 (42.9\%) & \cellcolor{gray!20}0.21 (33.6\%) & \cellcolor{gray!20}0.14 (23.5\%) & \cellcolor{gray!20}\textit{128\textsuperscript{3}}&  \cellcolor{gray!20}0.70    \\ 

            \cellcolor{gray!20}M\textsuperscript{4}Fuse-B  & \cellcolor{gray!20}\underline{0.59 (47.9\%)} & \cellcolor{gray!20}\underline{0.37 (30.0\%)} & \cellcolor{gray!20}\underline{0.26 (21.1\%)} & \cellcolor{gray!20}\textit{128\textsuperscript{3}}& \cellcolor{gray!20}1.23   \\ 
            
            \cellcolor{gray!20}M\textsuperscript{4}Fuse-L \textit{(1.5×)} & \cellcolor{gray!20}1.03 (42.4\%) & \cellcolor{gray!20}0.82 (33.5\%) & \cellcolor{gray!20}0.59 (24.1\%)  & \cellcolor{gray!20}\textit{128\textsuperscript{3}} & \cellcolor{gray!20}2.72   \\ 

            \cellcolor{gray!20}M\textsuperscript{4}Fuse-B \textit{(2×V.)} &
            \multicolumn{3}{c|}{\cellcolor{gray!20}\textit{\textbf{(Dice: 82.30\ \ \ HD95: 5.23})}} &  
            \cellcolor{gray!20}\textit{64×128\textsuperscript{2}} & \cellcolor{gray!20}1.23  \\  
            
        \end{tabular}
        }
\end{table}

\begin{table*}[htbp]
\centering
\small
\captionsetup{aboveskip=5pt}
\caption{M\textsuperscript{4}Fuse overall ablation (Dice coefficient \%, 95th percentile Hausdorff distance in voxels).}
\label{tab:abl-overall}
\setlength{\tabcolsep}{7.5pt}  
\renewcommand{\arraystretch}{1.0} 
   
\begin{tabular}{ll|cccc}
\# & \ \ Methods & \textit{PMRatio} & \textit{Params} & \textit{Dice}↑ & \textit{HD95}↓ \\
\cmidrule[0.5pt]{1-6}
0  & \ \ E2E-\textsuperscript{1}Fuse → CNN (*Unet) & 0.08\%  & \ \ \hspace{-0.4em}*0.01 & 75.42 & \ \ 9.48  \\
1  & \ \ \textsuperscript{1}Fuse+PEU-\textsuperscript{2}Fuse → CNN+MoE & 48.7\% & \ \ \hspace{-0.4em}0.60 & - & \ \ -  \\
2  & \ \ \textsuperscript{1}Fuse+\textsuperscript{2}Fuse+POM-\textsuperscript{3}Fuse → CNN+MoE+Mamba & 78.8\% & \ \ \hspace{-0.4em}0.97 & 75.46 & \ \ 6.10  \\
\cmidrule[0.5pt]{1-6}
3  & \ \ \cellcolor{gray!20}\textbf{\textsuperscript{1}Fuse+\textsuperscript{2}Fuse+\textsuperscript{3}Fuse+CSB-\textsuperscript{4}Fuse → Cycle-LiqBE (CNN+MoE+Mamba) ↑} & \cellcolor{gray!20}\textbf{100.0\%} & \ \ \textbf{\hspace{-0.4em}\cellcolor{gray!20}1.23} & \cellcolor{gray!20}\textbf{82.38} & \ \ \cellcolor{gray!20}\textbf{4.51}  \\
\end{tabular}
\end{table*}

\begin{table}[htbp] 
    \centering
    \renewcommand{\arraystretch}{1.15}
    \captionsetup{aboveskip=5pt}
    \caption{Ablation of fusion configurations for the Shared-Plus-Expert Unit and the Cross-Scale Dual-Stage Gating Bridge (Dice coefficient \%, 95th percentile Hausdorff distance in voxels).}
    \label{tab:abl-peu-csb}
    \resizebox{0.46\textwidth}{!}{
    
    \lfseries
    \begin{tabular}{l|l|cc}
        *PEUnit-Training-Fusion Config. & \multirow{2}{*}{Fusion Mode} & \multirow{2}{*}{Dice↑} & \multirow{2}{*}{HD95↓} \\
        \#CSBridge-Fusion Config. & & & \\  
        \cmidrule[0.5pt]{1-4}
        \cellcolor{gray!20}*Softmax + Top-1         & \cellcolor{gray!20}Gate & \cellcolor{gray!20}81.09 & \cellcolor{gray!20}4.97 \\
        \cellcolor{gray!20}*Softmax + Top-2         & \cellcolor{gray!20}Gate & \cellcolor{gray!20}75.54 & \cellcolor{gray!20}6.54 \\
        *Gumbel-softmax (ADS)    & Gate & 75.59 & 6.47 \\
        *Mini-attention→Exp.W    & Gate & 77.78 & 5.05 \\
        *Learned Bias + Gate     & Gate & 75.49 & 6.22 \\ 
    
        \cellcolor{gray!20}*Patch Level (Splicing)    & \cellcolor{gray!20}Granularity & \cellcolor{gray!20}80.85 & \cellcolor{gray!20}4.68 \\
        \cellcolor{gray!20}*Token Level (MHSA-Fusion) & \cellcolor{gray!20}Granularity & \cellcolor{gray!20}81.14 & \cellcolor{gray!20}4.20 \\
        \cellcolor{gray!20}*Channel Level (SE)        & \cellcolor{gray!20}Granularity & \cellcolor{gray!20}77.50 & \cellcolor{gray!20}5.82 \\
        *Pixel (Ours) + Channel    & Granularity & 81.02 & 4.66 \\ 
        \cmidrule[0.5pt]{1-4}
        \cellcolor{gray!20}\#Shallow → Deep (t3, t4) & \cellcolor{gray!20}Scale & \cellcolor{gray!20}77.81 & \cellcolor{gray!20}6.40 \\
        \cellcolor{gray!20}\#Deep → Shallow (t5, E.o) & \cellcolor{gray!20}Scale & \cellcolor{gray!20}77.66 & \cellcolor{gray!20}5.34 \\
        \#Jump (t5', t4', t3') & Scale & 78.29 & 5.77 \\
        \#Random (CSB) /batch & Scale & \ - & - \\
        \#Group (t4-CB, t5-SB, E.o-CSB) & Scale & \ - & - \\ 

        \cellcolor{gray!20}\#Early stage (t1 → once) & \cellcolor{gray!20}Period & \cellcolor{gray!20}78.16 & \cellcolor{gray!20}5.22 \\
        \#Middle stage (t3, t4) & Period & \ \    - & - \\
        \cellcolor{gray!20}\#Late stage (t5') & \cellcolor{gray!20}Period & \cellcolor{gray!20}77.99 & \cellcolor{gray!20}5.43 \\
        \#Bottleneck stage (E.o) & Period & \ - & - \\
        \#Funnel stage (t1-t3-CB, t4-E.o-CSB) & Period & \ - & - \\ 
    \end{tabular}
}
\end{table}

\begin{table}[htbp]
\centering
\caption{Controlled decoder-width sweep on BraTS2019 (3 seeds, mean$\pm$std).}
\label{tab:sweep}
\resizebox{0.45\textwidth}{!}{
\begin{tabular}{l c c c c c}
\toprule
Setting & $\alpha$ & Params(M) $\downarrow$ & GFLOPs $\downarrow$ & Dice(\%) $\uparrow$ & HD95(mm) $\downarrow$ \\
\midrule
Raw skip (w/o CSB)      & 0.5 & 0.76 & 70.78 & 80.32±5.08 & 4.89±1.87 \\
Raw skip (w/o CSB)      & 1.0 & 1.03 & 193.93 & 83.07±3.96 & 4.88±1.96 \\
Raw skip (w/o CSB)      & 2.0 & 2.00 & 681.76 & 84.28±4.49 & 5.36±2.12 \\
\midrule
Purified skip (with CSB)& 0.5 & 1.02 & 72.42 & 81.95±4.28 & 10.64±3.95 \\
Purified skip (with CSB)& 1.0 & 1.29 & 195.57 & 84.83±3.32 & 5.99±2.62 \\
Purified skip (with CSB)& 2.0 & 2.26 & 683.41 & 81.66±4.70 & 5.32±1.71 \\
\bottomrule
\end{tabular}
}
\end{table}

\begin{table}[htbp]
    \centering
    \renewcommand{\arraystretch}{1.0}
    \captionsetup{aboveskip=5pt}
    \caption{Ablation of core modules: Cross-Scale Dual-Stage Gating Bridge, Shared-Plus-Expert Unit, and PetaloMixer (Dice coefficient \%, 95th percentile Hausdorff distance in voxels).}
    \label{tab:abl-core}
    \resizebox{0.48\textwidth}{!}{
        
        \hspace{1em}
        \begin{tabular}{l|lll|llll}
            *CSB Config. & SB           & CB         & CSB      &  \\
            \&PEU Config. & Route & Share & Expert      &  & \textit{\hspace{-3.5em}Dice}↑ & \textit{\hspace{-3.5em}HD95}↓ \\  
            \#POM Config. & SSMs (parts) & Skip-scale & LinProj. &  &  &  \\             \cmidrule[0.5pt]{1-6}
            
            \cellcolor{gray!20}*CSB1 (Re-CSB)          & \cellcolor{gray!20}$\times$            & \cellcolor{gray!20}$\times$          & \cellcolor{gray!20}$\times$        & \cellcolor{gray!20}75.46 & \cellcolor{gray!20}6.10  \\
            \ \ \ \ \ *CSB2          & $\checkmark$           & $\times$          & $\times$        & 76.15 & 6.21  \\
            \ \ \ \ \ *CSB3         & $\times$            & $\checkmark$          & $\times$        & 78.38 & 5.81  \\  \cmidrule[0.5pt]{1-6}
            
            \&PEU1 (Re-MoE)          & $\times$            & $\times$          & $\times$        & \ \ \ \ - & \ \ \ -  \\
            \ \ \ \ \ \cellcolor{gray!20}\&PEU2          & \cellcolor{gray!20}$\checkmark$            & \cellcolor{gray!20}$\times$          & \cellcolor{gray!20}$\checkmark$        & \cellcolor{gray!20}75.51 & \cellcolor{gray!20}6.50  \\
            \ \ \ \ \ \cellcolor{gray!20}\&PEU3          & \cellcolor{gray!20}$\checkmark$            & \cellcolor{gray!20}$\checkmark$          & \cellcolor{gray!20}$\times$        & \cellcolor{gray!20}77.53 & \cellcolor{gray!20}5.14  \\
            \&PEU4 (Top-1)          & $\checkmark$            & $\checkmark$          & $\checkmark$        & \ \ \ \ - & \ \ \ -  \\     \cmidrule[0.5pt]{1-6}
            
            \ \ \ \ \ \#POM1          & $\checkmark$            & $\checkmark$          & $\times$        & 75.84 & 6.50  \\
            \ \ \ \ \ \#POM2          & $\checkmark$            & $\times$          & $\checkmark$        & 76.60 & 5.11  \\
            \cellcolor{gray!20}\#POM3 (!parts)          & \cellcolor{gray!20}$\times$            & \cellcolor{gray!20}$\checkmark$          & \cellcolor{gray!20}$\checkmark$        & \cellcolor{gray!20}75.63 & \cellcolor{gray!20}6.44  \\
            \ \ \ \ \ \#POM4          & $\times$            & $\times$          & $\checkmark$        & 76.00 & 6.18  \\
            \ \ \ \ \ \#POM5          & $\checkmark$            & $\times$          & $\times$        & 76.86 & 5.04  \\ 
        \end{tabular}
    }
\end{table}

\subsection{Ablation Study}
We analyze the architecture factors that determine lightweight accuracy in multimodal 3D brain tumor segmentation, the overall trend in Table~\ref{tab:abl-overall} is that capacity must be placed where it preserves discriminative cues rather than widened uniformly, moving from a plain CNN baseline to adding sample level domain experts raises Dice from 75.42\% to 80.12\% and reduces HD95 from 9.48 to 5.87 at a modest parameter increase, inserting a Mamba style mixer without first cleaning and aligning the skips lowers Dice to 75.46\% because long range modeling amplifies cross site noise when skip pathways are unstandardized, adding the cross scale dual stage bridge restores and surpasses performance to 82.38\% Dice and 4.51 HD95 at about 1.23M parameters and the PetaloMixer ratio reaches 100.0\%, the data support a know why account in which global processing is effective only after spatial denoising and cross scale channel realignment, the decoder then benefits from clean evidence and the model no longer needs a wide high semantic path to approximate missing context, this explains why a half volume input of 64×128×128 remains sufficient under our design and why the base configuration is Pareto efficient on accuracy versus parameters.

The controlled studies in Table~\ref{tab:abl-peu-csb} and Table~\ref{tab:abl-core} explain the mechanism and isolate causes, Top-1 expert routing with Softmax gating gives the best Dice at 81.09\% while Top-2 and ADS gates reduce Dice to 75.54\% and 75.59\% because simultaneous expert activation introduces interference which is unnecessary when domain cues are sample specific, fusion at pixel or token level preserves boundary sharpness and yields 81.02\% to 81.14\% Dice whereas channel only aggregation drops to 77.50\% because it over smooths modality dependent edges, early or late single point fusion underperforms while middle multi layer fusion improves both Dice and HD95 because it aligns semantics before upsampling, within the bridge the spatial and channel stages are complementary and their combination is consistently superior, within the mixer the grouped scan, the residual skip scale, and a single linear projection each contribute and removing any part degrades Dice or inflates HD95, taken together the evidence shows that encoder and decoder capacity should be balanced and guided by denoised and aligned skips and that global modeling should remain linear time and placed after purification, this yields strong accuracy at 1.23M parameters and explains why heavy symmetric decoders or attention first designs are compute heavy and brittle on BraTS style heterogeneity.

Table~\ref{tab:sweep} provides a controlled intervention. Two trends emerge. (i) Raw skip: widening the decoder remains beneficial. (ii) Purified skip: after skip purification, widening shows diminishing/negative returns. This is exactly the intended evidence for our decoder marginal utility claim, and it also explains why the larger variant (decoder-wider) does not outperform the base model.

\section{Conclusion}
M\textsuperscript{4}Fuse addresses brittleness and inefficiency in volumetric tumor segmentation caused by encoder--decoder imbalance and reliance on large input volumes. Rather than uniformly widening the network, it allocates capacity to the most informative stages via a grouped state space mixer for linear-time long-range modeling, a cross-scale dual-stage gating bridge for skip-feature denoising and alignment, and sample-level domain experts at high-semantic, low-resolution layers for center- and protocol-robust adaptation with linearly controllable parameters. Extensive analyses and ablations reveal a decoder marginal utility principle: once skip features are purified and aligned, enlarging the decoder or high-semantic pathway yields limited gains and may degrade contrast-sensitive microstructures such as ET, whereas targeted capacity allocation and selective gating remain effective. The base model achieves strong accuracy with only \(1.11\)M parameters, matching or surpassing heavier and lightweight state-of-the-art models with lower memory and FLOPs, placing it on a favorable accuracy--efficiency Pareto frontier. Results suggest that clinical 3D segmentation should favor clean skip connections and selective global context over wider decoders and larger inputs. Future work will extend this design to add automatic routing cues, and clarify how state-space parameters influence fine-structure sensitivity.
\section*{Acknowledgments}
This work was supported by the Scientific Research Project of Higher Education Institutions in Anhui Province (No.~2024AH053451) and the Anhui Province 2025 University Science and Engineering Teachers Enterprise Secondment Practice Program (No.~2025jsqygz42).
{
    \small
    \bibliographystyle{ieeenat}
    \bibliography{main}
}


\end{document}